\pdfoutput=1
\documentclass[10pt,twocolumn,letterpaper]{article}
\usepackage[utf8]{inputenc}
\usepackage{color}
\usepackage{graphicx}
\usepackage{booktabs}
\usepackage{multirow}
\usepackage[table,xcdraw]{xcolor}
\usepackage{authblk}
\usepackage{wacv}
\usepackage{times}
\usepackage{epsfig}
\usepackage{graphicx}
\usepackage{amsmath}
\usepackage{amssymb}
\usepackage{caption}
\def\wacvPaperID{601} 

\wacvfinaltrue

\title{Tool Detection and Operative Skill Assessment in Surgical Videos Using Region-Based Convolutional Neural Networks}

\newcommand*{\affaddr}[1]{#1} 

\begin{document}

\author{%
\textbf{Amy Jin, Serena Yeung, Jeffrey Jopling, Jonathan Krause, Dan Azagury,} \\ 
\vspace{-2ex}
\textbf{Arnold Milstein, and Li Fei-Fei}\\
{Stanford University}\\
\affaddr{Stanford, CA 94305}\\
}

\maketitle


\vspace{2ex}

\begin{abstract}

    Five billion people in the world lack access to quality surgical care. Surgeon skill varies dramatically, and many surgical patients suffer complications and avoidable harm. Improving surgical training and feedback would help to reduce the rate of complications—half of which have been shown to be preventable. To do this, it is essential to assess operative skill, a process that currently requires experts and is manual, time consuming, and subjective. In this work, we introduce an approach to automatically assess surgeon performance by tracking and analyzing tool movements in surgical videos, leveraging region-based convolutional neural networks. In order to study this problem, we also introduce a new dataset, m2cai16-tool-locations, which extends the m2cai16-tool dataset with spatial bounds of tools. While previous methods have addressed tool presence detection, ours is the first to not only detect presence but also spatially localize surgical tools in real-world laparoscopic surgical videos. We show that our method both effectively detects the spatial bounds of tools as well as significantly outperforms existing methods on tool presence detection. We further demonstrate the ability of our method to assess surgical quality through analysis of tool usage patterns, movement range, and economy of motion.
    
\end{abstract}

\section{Introduction}

Five billion people in the world lack access to safe surgical care \cite{alkire2015global}. According to the World Health Organization, there is a global mortality rate of 0.5-5\% for major procedures, and up to 25\% of patients undergoing operations that require a stay in the hospital suffer complications \cite{weiser2008estimation}. Many of these complications are caused by poor individual and team performance, for which inadequate training and feedback play an important role \cite{gawande1999incidence,healey2002complications,kable2002adverse}. These outcomes can be improved, as studies show that half of all adverse surgical events are preventable.\par

\begin{figure}
    \centering
    \includegraphics[scale=0.36]{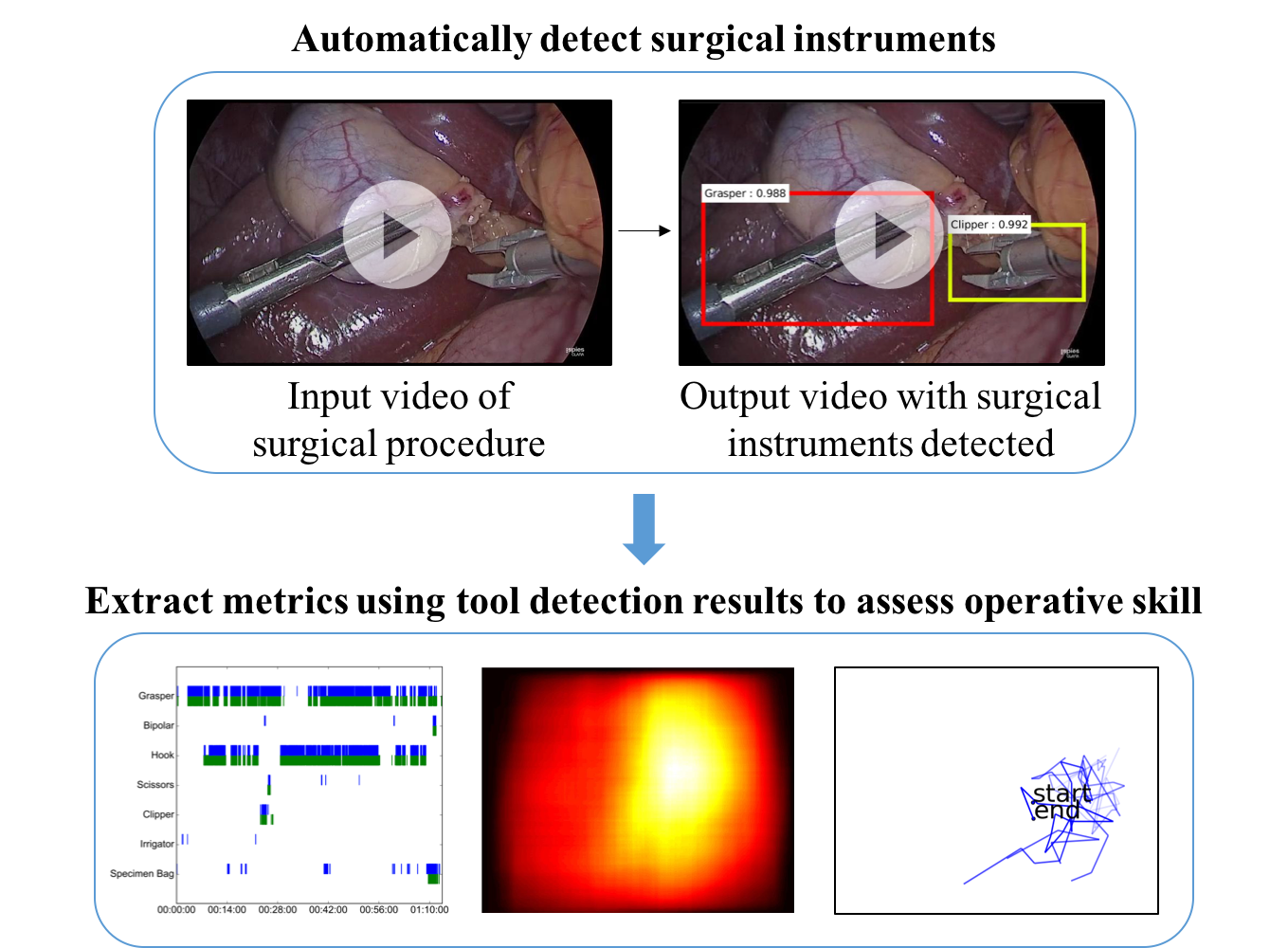}\\
    \caption{\small{Real-time automated surgical video analysis could facilitate objective and efficient assessment of surgical skill and provide feedback on surgical performance. In this work, we introduce an approach to: (1) automatically detect, classify, and localize surgical instruments in real-world laparoscopic surgical videos; and (2) efficiently analyze surgical performance based on the extracted tool information.}}
    \label{fig:overview}
\end{figure}

One major challenge to improving these outcomes is that surgeons currently lack individualized, objective feedback on their surgical technique and how to improve it. Manual assessment of surgeon performance requires expert supervision which is both subjective and time consuming, with many operations lasting hours. Real time automated surgical video analysis could provide a way to objectively and efficiently assess surgical skill.

A number of works address frame-level tool detection in laparoscopic surgical videos, including~\cite{sahu2016m2cai,twinanda2016single,raju2016m2cai,twinanda2016endonet}, as a part of the 2016 M2CAI Tool Presence Detection Challenge which includes a m2cai16-tool dataset \cite{m2cai}. However, rich analysis of surgeon performance involves analyzing tool movement such as movement range and economy of motion, and requires detecting the spatial bounds of tools in addition to their presence.

In this work, we address this task of spatial tool detection in laparoscopic surgical videos, which to our knowledge has not been previously studied. In order to study this, we introduce a new dataset, m2cai16-tool-locations, which extends the m2cai16-tool dataset with spatial bounds of tools. We develop an approach leveraging region-based convolutional neural networks (R-CNNs) to perform spatial detection of tools, and show that our method is able to both effectively detect the spatial bounds of tools as well as significantly outperform previous work on detecting tool presence.

Finally, because our deep learning approach allows for tool localization in addition to frame-level detection, it enables richer analysis of tool movements. We demonstrate the ability of our method to assess surgical quality through analysis of tool usage patterns, movement range, and economy of motion (Figure \ref{fig:overview}). We extract key quantitative and qualitative metrics that are proven to reflect surgical skill \cite{stylopoulosu2003celts}. We collaborated with surgeons who manually reviewed each surgical video. Their assessments substantiated our findings, validating the ability of our method to efficiently and accurately assess surgical quality. This enables not only avoiding the subjectivity inherent to human assessment, but also significantly reduces the time it takes to analyze a procedure.

In summary, we introduce m2cai16-tool-locations, a new dataset extending the m2cai16-tool dataset with spatial bounds of the tools. We use an approach leveraging region-based convolutional neural networks to effectively perform spatial detection of the tools, and show that conversion of these detections to frame-level presence detections also significantly outperforms state-of-the-art on that task. Importantly, we demonstrate that our spatial detections in these real-world laparoscopic surgical videos enables automatic assessment of surgical quality through analysis of tool usage patterns, movement range, and economy of motion.

\section{Related Work}

Early work on surgical tool detection, categorization, and tracking include those based on radio frequency identification (RFID) tags~\cite{kranzfelder2013rfid}; segmentation, contour processing and 3D modelling~\cite{speidel2009automatic}; and the Viola-Jones detection framework~\cite{lalys2012framework}.

Furthermore, deep learning approaches based on convolutional neural networks have shown impressive performance on computer vision tasks~\cite{russakovsky2015imagenet}, and works including \cite{sahu2016m2cai,twinanda2016single,raju2016m2cai,twinanda2016endonet} leverage deep learning architectures to achieve state-of-the-art performance on surgical tool presence detection and phase recognition. As a part of the M2CAI 2016 Tool Presence Detection Challenge~\cite{m2cai}, they introduced a benchmark for surgical tool presence detection.

While several existing studies address frame-level presence detection, Sarikaya \textit{et al.} perform surgical tool localization in videos of robot-assisted surgical training tasks, using multimodal convolutional neural networks~\cite{sarikaya2017}. Automated surgical scene understanding and skill assessment are further areas of study. A few studies have addressed specific components of surgical video analysis, including surgical phase recognition and activity recognition. As a part of the M2CAI 2016 Surgical Workflow Challenge, works including ~\cite{jin2016workflow, cadene2016workflow, twinanda2016endonet, stauder2017workflow} address surgical phase recognition in cholecystectomy videos. Additionally, Zia \textit{et al.} analyze task-specific suturing and knot-tying videos, using symbolic, texture, and frequency features \cite{zia2016}. Similarly, Lalys \textit{et al.} propose a framework using Hidden Markov Model and visual features, such as shape, color, and texture, to identify surgical tasks ~\cite{lalys2012framework}. Lea \textit{et al.} use skip-chain conditional random field as well as handcrafted features to classify actions in short segments of robotic surgery training tasks, including suturing and knot tying~\cite{lea2015}, and Reiter \textit{et al.} perform surgical tool localization and pose estimation; however, their approach is limited to robotic arms that return kinematic data \cite{reiter}. Also, Lin \textit{et al.} demonstrate the effectiveness of a Bayes classifier in differentiating the skill level of intermediate and expert surgeons on a suturing task, processing features extracted from motion data output by the da Vinci surgical system ~\cite{lin2010}. Additionally, Reiley \textit{et al.} use Hidden Markov Models for the task and gesture levels to assess robotic laparoscopic surgery ~\cite{reiley2009}. A limitation to these task-specific studies is that surgical tasks and gestures differ substantially from and do not accurately reflect surgical performance in real-world surgeries.

Our work builds on these prior contributions and uses region-based convolutional neural networks~\cite{ren2015faster} to detect the spatial bounds of tools, enabling richer, more comprehensive assessment of surgical quality in real-world laparoscopic cholecystectomies, or minimally invasive surgical removal of the gallbladder. Moreover, while studies often only use short segments of procedures or of surgeons performing simulated training tasks to analyze surgical skill, our work leverages unedited, full-length surgical operations. This differentiation is crucial for performance assessment, since real-time operations include smoke, lens fogging, variable anatomy, and different usage patterns not found in simulation scenarios. Limited segments of real operations may not give a comprehensive assessment of surgical performance. Thus, both simulation and video segments of actual operations are limited in their ability to facilitate assessment of surgical performance. In contrast, our extracted metrics are generated from comprehensive post-operative assessment of full surgical procedures.
They have the added benefit of being able to be correlated with post-operative clinical results, thus providing the link between surgical skill and outcomes.

\section{Dataset} 

\begin{figure*}
    \centering
    \includegraphics[scale=0.375]{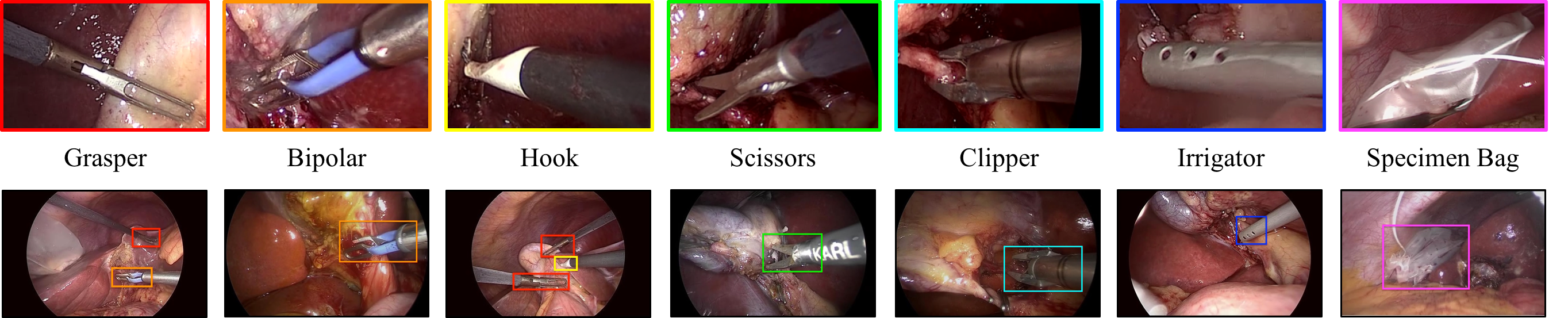}\\\centering
    \vspace{2ex}
    \captionsetup{justification=centering}
    \caption{\small{Top: the seven tools in m2cai16-tool-locations. Bottom: example frames with their spatial tool annotations. Color of the bounding box corresponds to tool identity.}}
    \label{fig:7tools}
\end{figure*}

There are few existing datasets on automated tool identification. Most center around frame-level tool presence detection, including m2cai16-tool, which was released for the M2CAI 2016 Tool Presence Detection Challenge, and Cholec80~\cite{twinanda2016endonet}. Both m2cai16-tool and Cholec80 contain cholecystectomy surgical videos performed at the University Hospital of Strasbourg in France, where each surgical video frame is labeled with binary annotations indicating tool presence. Sarikaya \textit{et al.} also address tool localization in videos of robot-assisted surgical training tasks ~\cite{sarikaya2017}. We expand upon these earlier works and introduce a new dataset to study the task of surgical tool localization in real-world laparoscopic surgeries and to enable higher-level analysis of surgical videos.

\begin{table}
\centering
\scalebox{0.84}{
\begin{tabular}{@{}cc@{}}
\toprule
\rowcolor[HTML]{EFEFEF} 
\textbf{Tool}         & \textbf{Number of annotated instances}   \\ \midrule
\textbf{Grasper}      & $923$          \\
\rowcolor[HTML]{EFEFEF} 
\textbf{Bipolar}      & $350$          \\
\textbf{Hook}         & $308$          \\
\rowcolor[HTML]{EFEFEF} 
\textbf{Scissors}     & $400$          \\
\textbf{Clipper}      & $400$          \\
\rowcolor[HTML]{EFEFEF} 
\textbf{Irrigator}    & $485$          \\
\textbf{Specimen Bag} & $275$          \\
\rowcolor[HTML]{EFEFEF} 
\textbf{Total}          & \textbf{$3141$} \\
\textbf{Number of Frames}          & \textbf{$2532$} \\
\bottomrule
\end{tabular}}
\caption{\small{Number of annotated frames for each tool.}}
\label{table:toolannot}
\end{table}

In order to study this task, a dataset containing annotations of spatial bounds of tools is required. However, to the best of our knowledge, no such dataset currently exists for real-world laparoscopic surgical videos. We therefore collect and introduce a new dataset, m2cai16-tool-locations, which extends the m2cai16-tool dataset \cite{m2cai} with spatial annotations of tools. We will publicly release this dataset.


m2cai16-tool consists of 15 videos recorded at 25 fps of cholecystectomy procedures. Videos 1 to 10 are used for training the R-CNN and videos 11 to 15 are used for testing the model. The videos, whose durations range from 20 to 75 minutes, are downsampled to 1 fps for processing. As a result, the dataset contains 23,000 frames labeled with binary annotations indicating presence or absence of seven surgical tools: grasper, bipolar, hook, scissors, clip applier, irrigator, and specimen bag.

In m2cai16-tool-locations, we label 2532 of the frames, under supervision and spot-checking from a surgeon, with the coordinates of spatial bounding boxes around the tools. We use 50\%, 30\%, and 20\% for training, validation, and test splits. The 2532 frames were selected from among the 23,000 total frames. We first annotate all frames containing just one tool, and increase the number of annotated instances per tool class by additionally labeling frames with two and three tools. The breakdown of number of annotations per tool class is detailed in Table~\ref{table:toolannot}. Figure \ref{fig:7tools} shows each tool in the dataset, its number of spatial annotations, and examples of annotations in a number of frames.

\section{Approach}


\begin{figure*}
    \centering
    \includegraphics[scale=0.47]{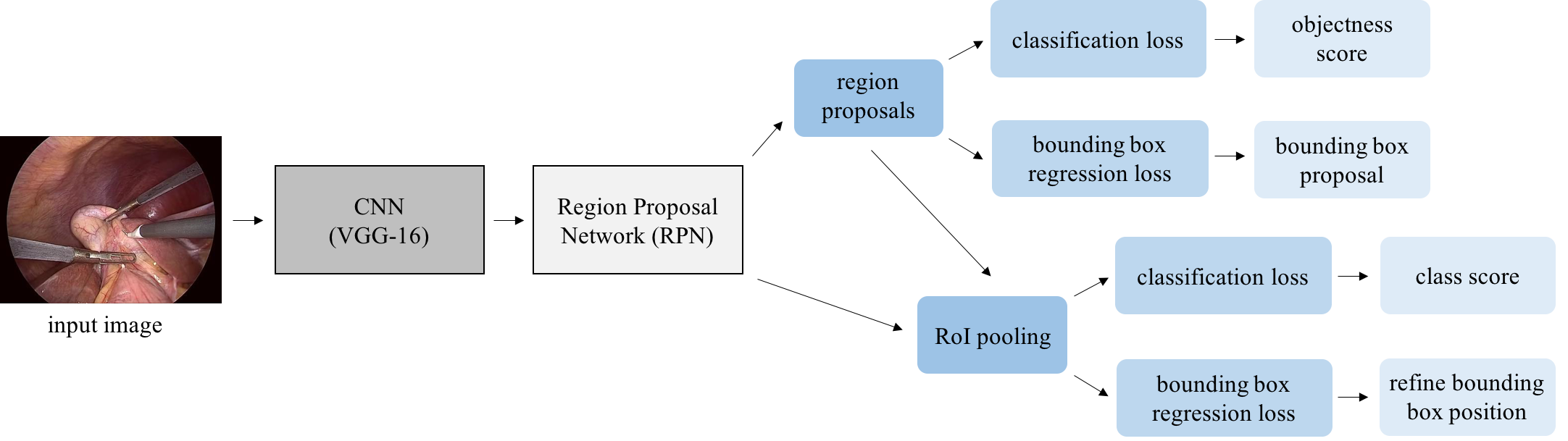}
    \captionsetup{justification=centering}
    \caption{\small{Faster R-CNN architecture. The input to the network is a frame from a surgical video. The base network of Faster R-CNN is a VGG-16 convolutional neural network. This is connected to a region proposal network (RPN) that shares convolutional features with object detection networks. For each input image, the RPN generates region proposals likely to contain an object, and features are pooled over these regions before being passed to a final classification and bounding box refinement network. The output is the spatial bounding box positions of detected surgical tools in the video frame.}}
    \label{fig:faster_rcnn}
\end{figure*}

Our approach for spatial detection of surgical tools is based on Faster R-CNN~\cite{ren2015faster}, a region-based convolutional neural network described in more detail below. The input is a video frame, and the output is the spatial coordinates of bounding boxes around any of the seven surgical instruments in Figure \ref{fig:7tools}. This output allows us to perform qualitative and quantitative analyses of tool movements, from tracking tool usage patterns to evaluating motion economy, and to correlate these measures with surgical skill, ultimately setting the stage for higher-level analysis of surgical performance.

The Faster R-CNN architecture we use is shown in Figure \ref{fig:faster_rcnn}. The base network is a VGG-16 convolutional neural network~\cite{simonyan2014very} with 16 convolutional layers, which extracts powerful visual features. On top of this network is a region proposal network (RPN) that shares convolutional features with object detection networks. For each input image, the RPN generates region proposals likely to contain an object, and features are pooled over these regions before being passed to a final classification and bounding box refinement network.  The use of the RPN enables significant computational gains over related previous work, R-CNN \cite{girshick2014rich} and Fast R-CNN \cite{girshick2015fast}.

The RPN is trained by optimizing the following loss function for each image:


\begin{equation}
\begin{split}
\label{eq1}
L(\{p_i\},\{t_i\}) & =\frac{1}{N_{cls}}\sum_{i}L_{cls}(p_i,p_i^*) \\ & +\lambda\frac{1}{N_{reg}}\sum_{i}p_i^*L_{reg}(t_i,t_i^*)
\end{split}
\end{equation}

Here \(i\) indexes ``anchors'' corresponding to each sliding window position of the input feature map, \(p_i\) is the anchor's objectness probability, and \(t_i\) is the coordinates of the predicted bounding box. \(p_i^*\) is the ground-truth label of whether an anchor is a true object location based on Intersection over Union (IoU) with ground-truth annotations, and \(t_i^*\) is the coordinates of the ground-truth box corresponding to a positive anchor. The loss function is therefore a weighted combination of a classification loss \(L_{cls}\) for the binary objectness label and a regression loss \(L_{reg}\) for bounding box coordinates. \(N_{cls}\) and \(N_{reg}\) are normalization constants and \(\lambda\) weights the contributions of classification and regression. The classification and bounding box refinement networks on top of the pooled regions of interest are trained using standard cross-entropy and regression loss functions.

While Faster R-CNN has shown impressive performance on detection of everyday objects, the domain of surgical videos and surgical tools has quite different visual characteristics.  We pre-train the network on the ImageNet dataset~\cite{russakovsky2015imagenet}, where a large amount of data is available to learn general visual features, and then fine-tune the network on our m2cai16-tool-locations dataset, where a smaller amount of data is labeled with the surgical tools of interest.

To train the RPN, we assign a binary objectness label to each anchor at each sliding window position of the feature map. We also assign a positive label to anchors with an overlap greater than 0.8 with the ground-truth box or if those do not exist, to an anchor or anchors with the highest Intersection over Union (IoU), and a negative label to anchors with an IoU of less than 0.3.

We fine-tune the VGG-16 network to optimize model performance using stochastic gradient descent. We modify the classification layer of the network to output softmax probabilities over the seven tools. All layers are fine-tuned for 40K iterations with a mini-batch size of 50, and a 3$\times$3 kernel size is used. We perform data augmentation by randomly flipping frames horizontally. The learning rate is initialized at 10$^{-3}$ for all layers, and decreased by a factor of 10 every 10K iterations. Total training time was approximately two days on an NVIDIA Tesla K40 GPU, and the network's processing speed at deployment is 5 fps, achieving real-time surgical tool detection.

\section{Results}

In this section, we quantitatively evaluate our approach on the tasks of spatial detection and frame-level presence detection of surgical tools, using m2cai16-tool-locations and m2cai16-tool, respectively. We demonstrate strong performance on the new task of spatial detection, and by leveraging spatial annotations, we significantly outperform existing works on presence detection. Finally, we qualitatively illustrate the ability of our approach to analyze tool usage patterns, movement range, and economy of motion for assessment of surgical performance.

\subsection{Spatial detection}

\begin{figure*}
    \centering
    \includegraphics[scale=0.5]{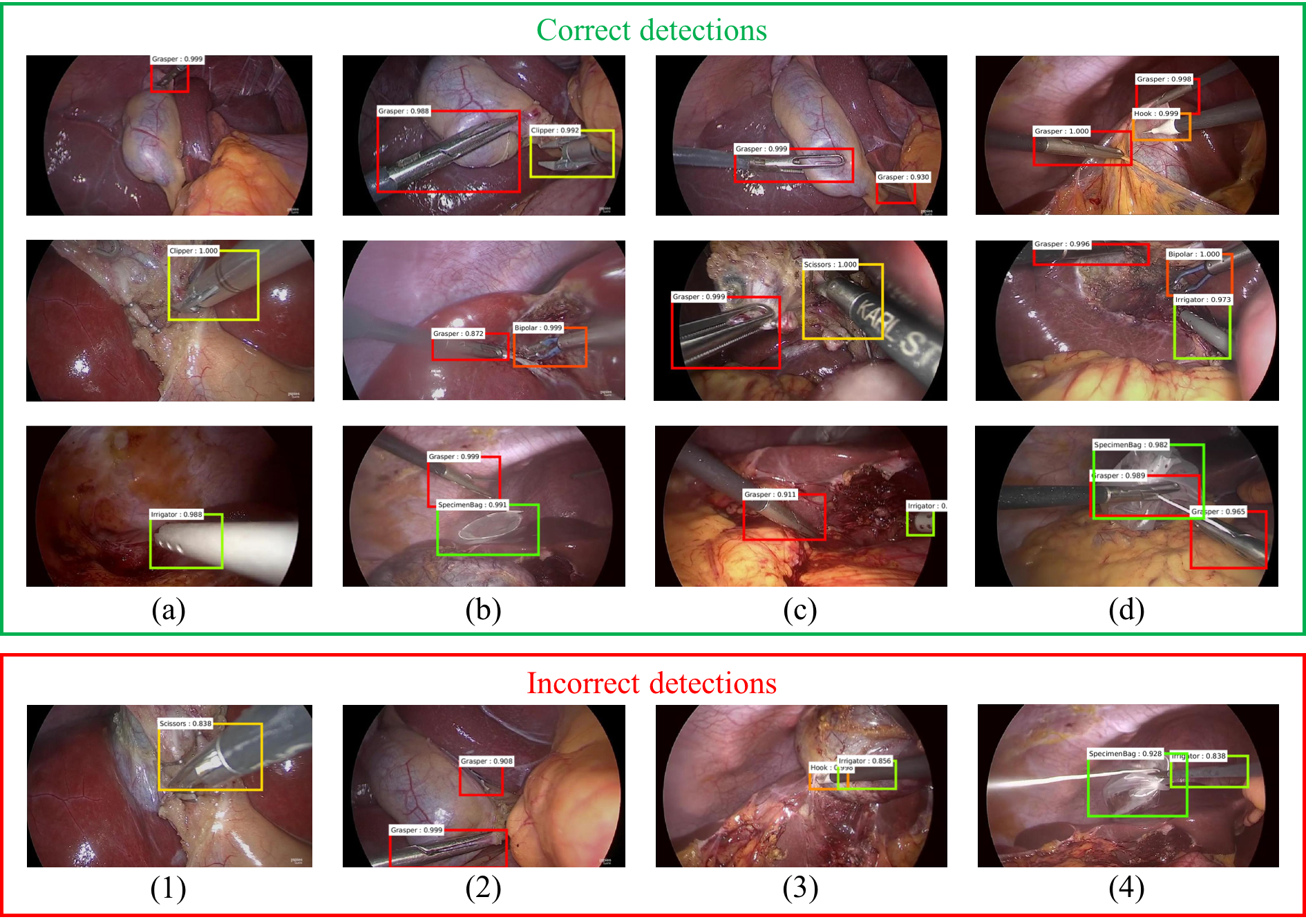}
    \captionsetup{justification=centering}
    \caption{\small{Example frames of spatial detection results. Bounding box color corresponds to predicted tool identity. Correct predictions are boxed in green (top), and mistakes are boxed in red (bottom). The model is able to successfully detect, classify, and localize surgical instruments despite varying tool positions and angles, and despite some parts of the tools being occluded, as shown in column (c).}}
    \label{fig:sample}
\end{figure*}

To the best of our knowledge, this study was the first to perform surgical tool localization in real-world laparoscopic surgical videos, which would set the stage for richer analysis of surgical performance. Table \ref{table:spatialAP} presents performance using average precision (AP) on spatial detection of the surgical tools in m2cai16-tool-locations.

\begin{table}[h!]
\centering
\scalebox{0.84}{
\begin{tabular}{@{}cc@{}}
\toprule
\rowcolor[HTML]{EFEFEF} 
\textbf{Tool}         & \textbf{AP}   \\ \midrule
\textbf{Grasper}      & $48.3$          \\
\rowcolor[HTML]{EFEFEF} 
\textbf{Bipolar}      & $67.0$          \\
\textbf{Hook}         & $78.4$          \\
\rowcolor[HTML]{EFEFEF} 
\textbf{Scissors}     & $67.7$          \\
\textbf{Clipper}      & $86.3$          \\
\rowcolor[HTML]{EFEFEF} 
\textbf{Irrigator}    & $17.5$          \\
\textbf{Specimen Bag} & $76.3$          \\
\rowcolor[HTML]{EFEFEF} 
\textbf{mAP}          & $\mathbf{63.1}$ \\ \bottomrule
\end{tabular}}
\caption{\small{Spatial detection average precision (AP) per-class and mean average precision (mAP) in m2cai16-tool-locations. Clipper achieves highest performance, likely due to its well-visualized usage pattern. Irrigator has notably low performance, likely due to its generic shape (it looks like the pole of every other instrument) and its lack of use in every procedure.}}
\label{table:spatialAP}
\end{table}

Intersection over Union (IoU) of 0.5 with a ground-truth bounding box for a given class is considered to be a correct detection. The overall mAP is 63.1, indicating strong performance overall. Clipper is the highest performing tool, likely due to its usage pattern: while only present for a specific step in cholecystectomies, surgeons make sure it is well visualized the entire time it is being used. Thus, its consistent usage pattern may have contributed to its high performance. The irrigator, on the other hand, is difficult likely due to its generic shape (it looks like the pole of every other instrument) and its lack of use in every procedure.

Figure~\ref{fig:sample} provides example frames of detection results. Columns (a) through (d) display frames with increasing numbers of tools present in them; from left to right, either 1, 2, or 3 tools are present per frame. We find that our model is able to successfully detect, classify, and localize surgical instruments despite varying tool positions and angles, and despite some parts of the tools being occluded, as shown in column (c). (1) through (4) present incorrect or partially incorrect detections. In (1), the clipper is mistaken as a scissor; perhaps the angle of the tool renders its identity ambiguous. In (2), the center grasper bounding box is a false positive; rather, the structure and shape of the gallbladder and liver form an outline that may appear to be a grasper to the model. Interestingly, in both (3) and (4), the poles of the hook and grasper, respectively, are identified as an irrigator. This mistake is not surprising, as the irrigator tip takes on the generic shape of any pole of a tool. Additionally, (4) presents a false negative, as the grasper in the center of the frame is left undetected by the model. Overall, our model has strong performance on the task of spatial tool detection.




\subsection{Frame-level presence detection}
The spatial detections output from our model can also be converted to frame-level presence predictions, and evaluated on the m2cai16-tool presence detection benchmark. Table~\ref{table:frameAP} presents AP performance on frame-level presence detection of tools in m2cai16-tool.

\begin{table}[h!]
\centering

\scalebox{0.84}{
\begin{tabular}{@{}cc@{}}
\toprule
\rowcolor[HTML]{EFEFEF} 
\textbf{Tool}         & \textbf{AP}   \\ \midrule
\textbf{Grasper}      & $87.2$          \\
\rowcolor[HTML]{EFEFEF} 
\textbf{Bipolar}      & $75.1$          \\
\textbf{Hook}         & $95.3$          \\
\rowcolor[HTML]{EFEFEF} 
\textbf{Scissors}     & $70.8$          \\
\textbf{Clipper}      & $88.4$          \\
\rowcolor[HTML]{EFEFEF} 
\textbf{Irrigator}    & $73.5$          \\
\textbf{Specimen Bag} & $82.1$          \\
\rowcolor[HTML]{EFEFEF} 
\textbf{mAP}          & $\mathbf{81.8}$ \\ \bottomrule
\end{tabular}}

\caption{\small{Frame-level presence detection average precision (AP) per-class and mean average precision (mAP) in m2cai16-tool. Hook achieves the highest performance, perhaps due to its distinct tip shape. Scissors has lowest performance, most likely due to its brief usage during the operation and common two-pronged tip shape.}}
\label{table:frameAP}
\end{table}

Hook achieves the highest performance; one possible explanation is that it has a distinct tip shape, making it easily distinguishable from other tools. Clipper, grasper, and specimen bag also perform well, while bipolar and irrigator are more frequently misidentified, perhaps due to their generic tip shapes as well as sparse and irregular appearances, as usage of these tools is not absolutely essential during the cholecystectomy procedure. For the bipolar and irrigator, usage is conditional on the real-time circumstances of the operation. The bipolar, an instrument for dissection and hemostasis, is used to prevent likely bleeding, or to cauterize bleeding when it does occur. The irrigator, an instrument used for flushing and suctioning an area, is typically used during a cholecystectomy procedure when blood accumulates in the surgical field. Scissors has lowest performance, most likely due to its brief usage during the operation and common two-pronged tip shape.

\begin{figure}
    \centering
    \includegraphics[scale=0.38]{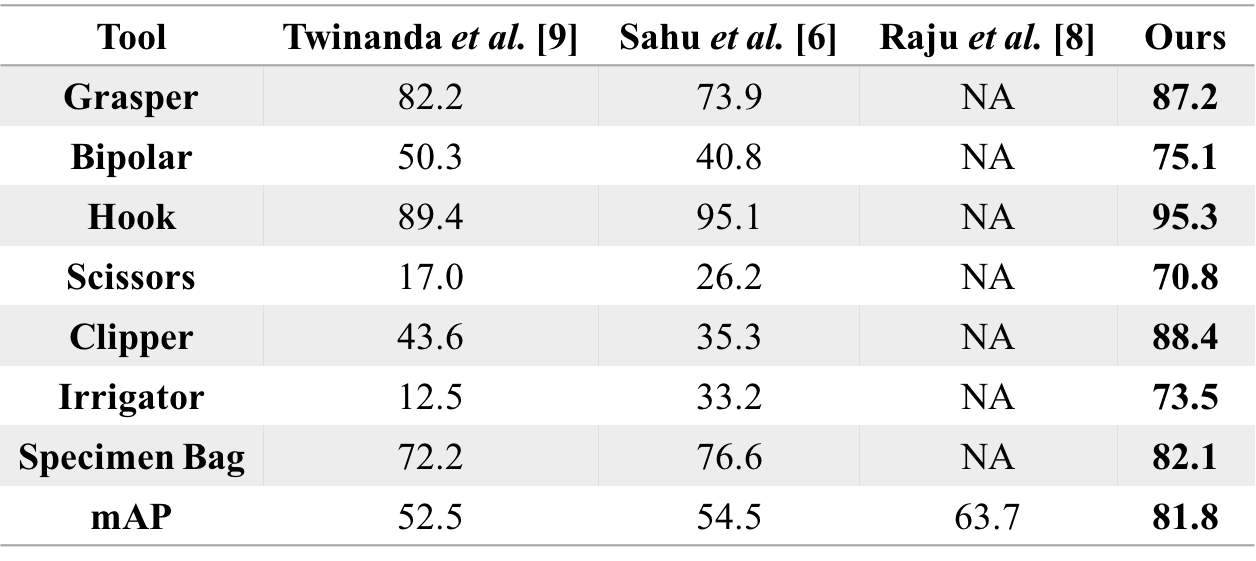}
    \captionsetup{justification=centering}
    \caption{\small{Comparison with winners of the M2CAI Tool Presence Detection Challenge~\cite{m2cai} on frame-level presence detection in m2cai16-tool. By leveraging the new spatial annotations during training, our approach is able to significantly outperform all previous work.}}
    \vspace{-4ex}
    \label{fig:comptable}
\end{figure}

Figure~\ref{fig:comptable} compares mAP performance with winners of the 2016 M2CAI Tool Presence Detection Challenge~\cite{m2cai}. By leveraging the new spatial annotations during training, our approach is able to significantly outperform all previous work. In comparison to the M2CAI 2016 Tool Presence Detection Challenge studies, we achieve a 28\% improvement in accuracy, increasing mAP from 63.8 to 81.8, using just 2,500 of the 23,000 frames—roughly a tenth of the data—in m2cai16-tool, which the previous studies used in its entirety to train their tool presence detection models. We also surpass EndoNet~\cite{twinanda2016endonet}, a novel CNN architecture based on AlexNet~\cite{krizhevsky2012alexnet}, in the tool presence detection task.

\subsection{Assessment of Surgical Performance}

\begin{figure*}[h!]
    \centering
    \includegraphics[scale=0.55]{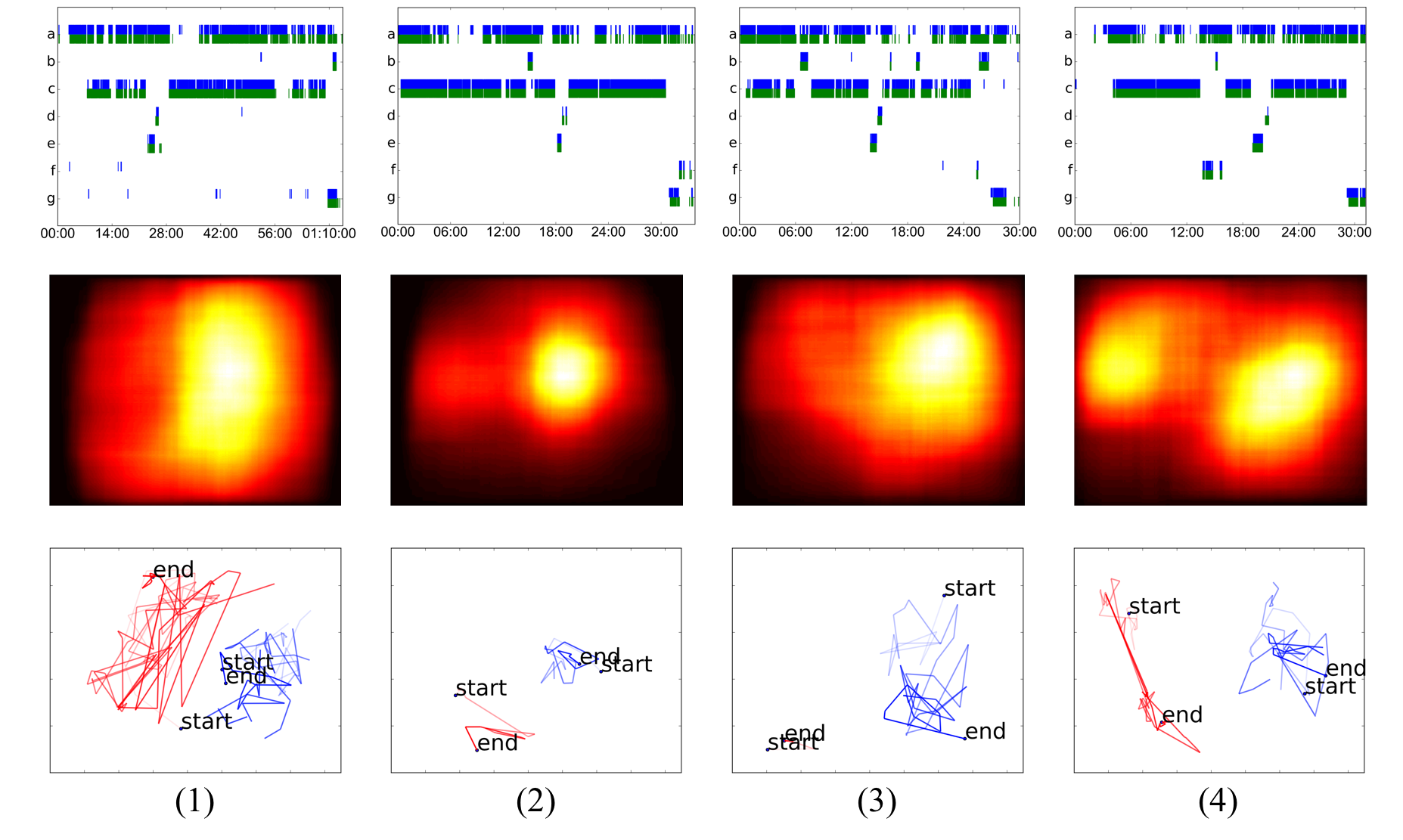}
    \caption{\small{Timelines (top), heat maps (middle), and trajectories (bottom) of tool usage for the testing videos 1 through 4 in m2cai16-tool. In the timelines, (a)-(g) correspond to Grasper, Bipolar, Hook, Scissors, Clipper, Irrigator, and Specimen Bag, respectively. These metrics effectively measure bimanual dexterity, efficiency, and overall operative skill and enable us to efficiently examine back and forth switching of instruments, movement range, and motion patterns of tools. We find that testing video 2 correlates with the most well-executed surgery, reflecting focused and skillful execution of each step of the surgical procedure. In contrast, the surgeons in the other testing videos have much less economy of motion, handle the instruments with less dexterity, and struggle with certain parts of the procedure.}}
    \vspace{-2ex}
    \label{fig:fig}
\end{figure*}

The current standard for surgical performance assessment is to have an expert surgeon observe the operation in its entirety and provide feedback to the surgeon performing the surgery. The Global Operative Assessment of Laparoscopic Skills (GOALS) rating system, is a validated rubric for grading surgeons on laparoscopic surgery performance. For each of the five assessment domains of GOALS—depth perception, bimanual dexterity, efficiency, and tissue handling, and autonomy, surgeons are rated on a scale of 1 to 5, from least to most technically proficient, totaling to a final score out of 25.

Operative skill significantly impacts patient outcomes. Birkmeyer \textit{et al.}~\cite{birkmeyer2013surgical} has demonstrated that performance on standardized ratings of surgical skill are correlated with patient complication rates and hospital readmissions. Despite this, most surgeons get no formal feedback on their operative performance. The process of manually rating surgeries is time-consuming and subject to bias. Motivated by these factors, we use the output of our deep learning model, which consists of frame-by-frame tool presence, identity, and location information, to extract key metrics that are proven to reflect surgical skill, such as instrument usage times and path length~\cite{stylopoulosu2003celts}. We also create visual depictions of the progression of the operation, like instrument usage timelines and tool trajectory maps, to parallel the GOALS rating system.

By working with a group of surgeons who each independently reviewed and rated each testing video manually, we are able to validate our approach to assess surgical performance on the five testing videos in our dataset. In the following paragraphs, we interweave the surgeons' independent ratings and comments with our objective analyses of the surgical performance in each video.

To assess instrument usage patterns, we generate timelines displaying tool usage over the course of each of the five testing videos (Figure \ref{fig:fig}, top). In examining the timelines, we find that testing video 2 correlates with the most well-executed surgery. The surgery itself is efficient, and the number of times that different instruments are switched out for one another is minimal, reflecting focused and skillful execution of each step of the surgical procedure. On the contrary, the timeline for testing video 3 reflects inefficient and poorer technique, with more frequent switching back and forth of instruments.

\begin{figure}
    \centering
    \includegraphics[scale=0.38]{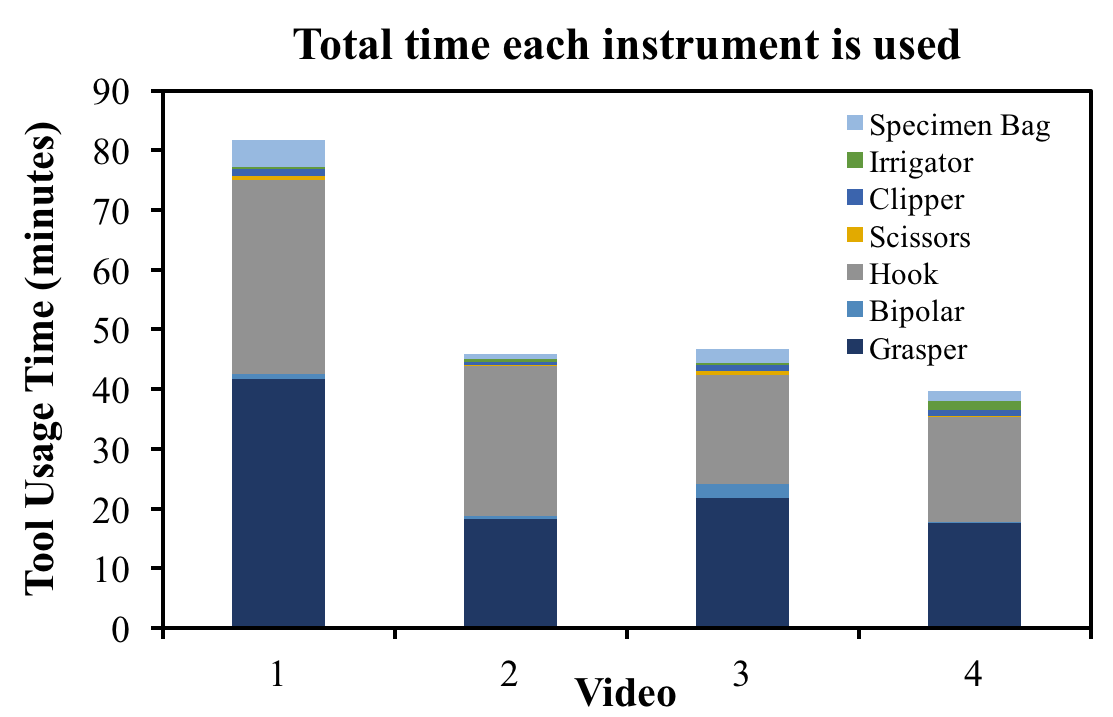}
    \caption{\small{Total instrument usage times, by video. Reflecting level of skill in handling tissue, the longer presence of the bipolar in testing video 3 indicates tissue damage, as the bipolar is used to stop bleeding.}}
    \vspace{-3ex}
    \label{fig:toolusagetime}
\end{figure}

Moreover, the instrument usage timelines and bar graphs quantifying the total time each instrument is used reflect the level of technical proficiency in handling tissue (Figure \ref{fig:toolusagetime}). The longer presence and periodic appearance of the bipolar in testing video 3 indicates rough tissue handling resulting in tissue damage, as the bipolar is used to stop bleeding.

\begin{figure}[h!]
    \centering
    \includegraphics[scale=0.38]{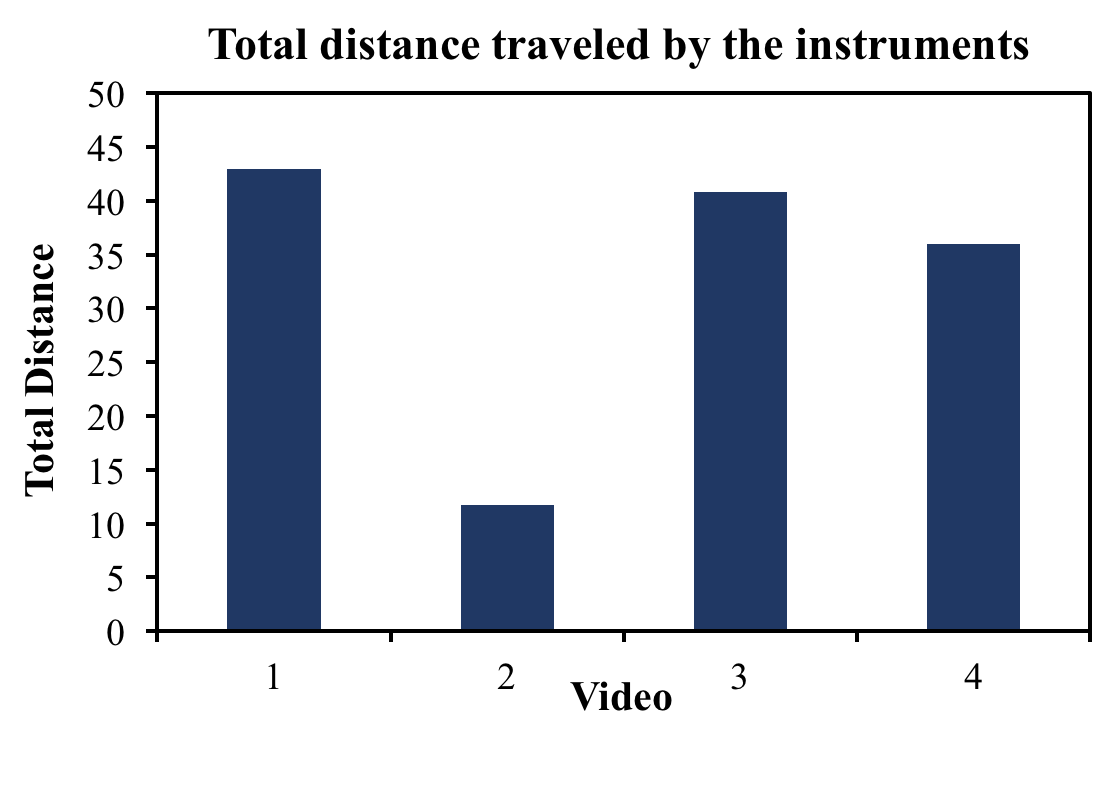}
    \caption{\small{Total distance traveled by the tools during the clipping phase, by video. Testing video 2 reflects the most technically excellent surgery, with the greatest economy of motion and most deft tool handling.}}
    \vspace{-3ex}
    \label{fig:totaldistance}
\end{figure}


To study the movement range of the surgical instruments, we generate heat maps of bounding box occurrences and locations. Figure \ref{fig:fig}, middle shows the heat maps of the testing videos. As an indicator of technical proficiency, especially of bimanual dexterity and efficiency, the heat maps reveal the surgical skill demonstrated in testing video 2 to be the highest among the test set once again. Better surgeons generally handle instruments in a more focused region of the operative field, exhibiting greater precision and economy of motion.



Finally, our deep learning model for automated tool detection and localization enabled tool tracking to study motion patterns. To further gain insight into bimanual dexterity and efficiency, we generated tool trajectory maps for the clipping phase of each of the five testing videos. We chose to examine this phase in particular because placing clips on the cystic artery and cystic duct is one of the most important steps of the cholecystectomy procedure. If these clips are placed in the wrong place, or if they come loose, the patient can suffer devastating complications.

The clipper (blue) and grasper (red) trajectories in testing video 2 reflect the deftness with which the surgeon placed the clips. The trajectory maps for testing videos 2 and 3 also provide valuable information for evaluating bimanual dexterity. While the clipper is used to clip the cystic artery and cystic duct, the grasper is used to hold the gallbladder in place and facilitate clip application. Based on the trajectory maps, the surgeons in testing videos 2 and 3 adeptly maneuver the grasper in a way such that they do not need to constantly adjust position, tension, or grip, in contrast to the trajectory maps for testing videos 1, 4, and 5. Indeed, when surgeons manually reviewed each of the testing videos to rate them with GOALS, they found that the surgeon in testing video 1 struggled to place the clips properly. He or she placed an additional clip, had difficulty prying it loose, and ultimately removed it.

We also quantitatively measured efficiency by computing total distance traveled by the tools during the clipping phase (Figure \ref{fig:totaldistance}). Once again, testing video 2 reflects the most technically excellent surgery, with the greatest economy of motion and most deft tool handling.



\subsection{Validation of approach with GOALS}
In addition to the blinded, subjective assessments of the test videos embedded in the preceding paragraphs, the three surgeons independently rated four of the test videos using a modified version of the GOALS assessment rubric. For this study the autonomy domain was omitted, which can only be assessed in person and is not applicable to every surgery. Thus, the total possible score for each video was 20. Test videos 1-4 received average composite scores of 10.00, 18.67, 9.33, and 13.33. The scores and subcomponent scores of GOALS, such as the efficiency score, directly correlated with the metrics extracted by our algorithm, such as the heat maps and tool trajectories. Table ~\ref{table:goals} presents the GOALS surgeon ratings for each of the testing videos. 

\begin{table}[]
\centering
\scalebox{0.84}{
\begin{tabular}{@{}ccccc@{}}
\toprule
\rowcolor[HTML]{EFEFEF} 
\textbf{}                   & \textbf{Video 1} & \textbf{Video 2} & \textbf{Video 3} & \textbf{Video 4} \\ \midrule
\textbf{Depth Perception}   & $2.67$             & $4.67$             & $2.33$             & $3.67$             \\
\rowcolor[HTML]{EFEFEF} 
\textbf{Bimanual Dexterity} & $3.00$             & $4.67$             & $2.00$             & $3.33$             \\
\rowcolor[HTML]{FFFFFF} 
\textbf{Efficiency}         & $2.00$             & $4.67$             & $2.33$             & $3.00$             \\
\rowcolor[HTML]{EFEFEF} 
\textbf{Tissue Handling}    & $2.33$             & $4.67$             & $2.67$             & $3.33$             \\
\rowcolor[HTML]{FFFFFF} 
\textbf{Total}              & \textbf{$10.00$}   & \textbf{$18.67$}   & \textbf{$9.33$}    & \textbf{$13.33$}   \\ \bottomrule
\end{tabular}}

\caption{\small{Average composite GOALS ratings for the four testing videos, including average subcomponent scores, as rated independently by the three surgeons. According to these ratings, the surgeon in testing video 2 is the most skilled, with a high score of 18.67 out of a maximum of 20.00, and the surgeons in testing videos 1 and 3 need to improve their operative technique, with scores of 10.00 and 9.33, respectively. These findings are consistent with our analyses based on the extracted assessment metrics, proving our approach to be a much more efficient way to review surgeon performance.}}
\label{table:goals}
\end{table}

\section{Conclusion}

In this work, we present a new dataset m2cai16-tool-locations and an approach based on region-based convolutional neural networks, to address the task of spatial tool detection in real-world laparoscopic surgical videos. We show that our method achieves strong performance on the new spatial detection task, outperforms previous work on frame-level presence detection, and runs in real-time at a frame rate of 5fps. Furthermore, it can be used to extract rich surgical assessment metrics such as tool usage patterns, movement range, and economy of motion, which directly correlate with independent assessments of the same videos made by experienced surgeons. Future work includes continuing to build on the types of meaningful information that can be automatically extracted from surgical videos, including smoothness of motion, tissue damage, repeated movements, and phase of procedure, as well as developing an automated GOALS rating system.




{\footnotesize 

}

\end{document}